# EVIDENCE ABSORPTION AND PROPAGATION THROUGH EVIDENCE REVERSALS

Ross D. Shachter
Department of Engineering-Economic Systems
Stanford University
Stanford, CA 94305-4025
shachter@sumex-aim.stanford.edu

The arc reversal/node reduction approach to probabilistic inference is extended to include the case of instantiated evidence by an operation called "evidence reversal." This not only provides a technique for computing posterior joint distributions on general belief networks, but also provides insight into the methods of Pearl [1986a] and Lauritzen and Spiegelhalter [1988]. Although it is well understood that the latter two algorithms are closely related, in fact all three algorithms are identical whenever the belief network is a forest.

## 1. Introduction

In recent years, three main classes of exact algorithms have emerged to solve probabilistic inference problems formulated as belief networks (or influence diagrams). The propagation method of Pearl [1986a] is an efficient, message-passing approach for polytrees (singly-connected networks, or directed graphs in which there are no undirected cycles), which can be generalized through conditioning to more general networks [Pearl 1986b]. The method of arc reversals and node reductions [Shachter 1986, 1988] processes general networks through topological transformations which preserve criteria values and joint distributions. The newest method was developed by Lauritzen and Spiegelhalter [1988] and generalized to Dempster-Shafer belief functions by Dempster and Kong [1986] and Shafer et al [1987]. It constructs a chordal undirected graph analog to the belief network in order to obtain processing efficiency on general networks similar to Pearl's method. It works on a special case of a polytree, a forest (a disconnected set of trees, or a directed acyclic graph in which no node has more than one parent).

In this paper, the method of arc reversals and node reductions is extended to efficiently handle experimental evidence. Originally developed for processing of influence diagrams with decisions [Howard and Matheson 1981], this reversal approach uses simple graph reductions to transform the topology of the network, while maintaining the joint distribution of a subset of the variables or the (expected) value of criterion variables [Olmsted 1983, Shachter 1986, 1988]. In decision analysis with sequential decisions, most of the experimental evidence is observed after the initial decision, so the emphasis in the method has been on pre-posterior analysis, that is, planning for all possible values of the experimental outcome. In this paper, however, special care is taken to efficiently process evidence which is observed prior to the initial decision. On a diagram without any decisions, this is precisely the probabilistic inference problem on belief networks.

Section 2 defines the belief diagram while Section 3 defines an evidence node and the operations of evidence absorption, reversal, and propagation. Section 4 introduces probability propagation, and the control of these operations is described in Section 5. Section 6 contrasts this method with those of Pearl and of Lauritzen and Spiegelhalter and Section 7 contains some conclusions.

## 2. Belief Diagrams

The necessary notation is presented in this section to define a the belief diagram, a generalization of a probabilistic influence diagram. Although the results in this paper can easily be applied to the general influence diagram with decision and value nodes, only chance nodes, representing random variables, will be used to simplify the presentation. This probabilistic influence diagram with evidence nodes corresponds exactly to belief networks, and from here on, they will be referred to as belief diagrams.

A belief diagram is a network built on a directed acyclic graph. The probabilistic nodes $N=\{1, \ldots, n\}$ correspond to random variables $X_1, \ldots, X_n$. Each variable $X_j$ has a set possible outcomes, $\Omega_j$, and a conditional probability distribution, $\pi_j$, over those outcomes. The conditioning variables for $\pi_j$ have indices in the set of parents or conditional predecessors, $C(j)$, $C(j) \subset N$, and are indicated in the graph by arcs from the nodes in $C(j)$ into node j. If $\pi_j$ is a marginal distribution, then $C(j)$ is the empty set, $\emptyset$.

Each probabilistic variable $X_j$ is initially unobserved, but at some time its value $x_j \in \Omega_j$ might become known. At that point, it becomes an evidence variable, as discussed in the next section.

As a convention, a lower case letter represents a single node in the graph and an upper case letter represents a set of nodes. If J is a set of nodes, $J \subseteq N$, then $X_J$ denotes the vector of variables indexed by J and $\Omega_J$ denotes the cross product of their outcomes, $\times_{j \in J}\Omega_j$. For example, the conditioning variables of $X_j$ are $X_{C(j)}$ and they have outcomes $\Omega_{C(j)}$.

The set of children or (direct) successors, $S(j)$, of the node $j$, is given by
$S(j) = \{ i \in N: j \in C(i) \}$.
It is also convenient to keep track of the ancestor nodes or indirect predecessors of node $j$, which are defined to include the parents of node $j$.
A list of nodes is called ordered if none of the indirect predecessors of a node follow the node in the list. Such a list exists if and only if there is no directed cycle among the nodes.

Given this notation there is a clear relation between the evidence diagram and the joint probability distribution. The conditional probability distributions $\{\pi_j\}$ are simply a factorization of the joint distribution $\Pr\{X_N\}$. If $\{1, \ldots, n\}$ is an ordered list of the nodes N in the diagram, then
$$\Pr\{X_N\}$$
$$= \Pr\{X_1\} \cdot \Pr\{X_2 | X_1\}$$
$$\cdot \ldots \cdot \Pr\{X_n | X_1, \ldots, X_{n-1}\}$$
$$= \pi_1(X_1 | X_{C(1)}) \cdot \pi_2(X_2 | X_{C(2)})$$
$$\cdot \ldots \cdot \pi_n(X_n | X_{C(n)}),$$
where $\pi_1$ is just the marginal distribution for $X_1$ since $C(1)$ is the null set, $\emptyset$. Since the joint distribution must sum to one, each of the conditional distributions could be multiplied by a constant without destroying any information. In that case,
$$\Pr\{X_N\}$$
$$\propto \pi_1(X_1 | X_{C(1)}) \cdot \pi_2(X_2 | X_{C(2)})$$
$$\cdot \ldots \cdot \pi_n(X_n | X_{C(n)}).$$
From here on, the distributions $\{\pi_j\}$ will be assumed to be defined only to within a constant of proportionality.

## 3. Evidence Nodes and Evidence Propagation

In this section evidence nodes are introduced to represent variables in the belief diagram whose values have been observed. The process of instantiating a variable's value will be called evidence absorption. This information is then spread through the network by a process called evidence propagation, which consists of repeated applications of evidence reversal, an operation which represents Bayes' Theorem in the belief diagram.

When a variable $X_j$ is observed with value $x_j$, the belief diagram must be modified to reflect the instantiation of the node j, in a way that maintains the posterior joint distribution. This operation, called evidence absorption, consists of modifications to the distributions (and outcomes) at node j and to the distributions stored in its successors. (Evidence absorption is closely related to the operation by that name in the Lauritzen and Spiegelhalter [1988] algorithm.) Once $X_j$ has been observed, it is said to be an evidence variable, and its corresponding node, called an evidence node, is drawn with shading. Since the probability of outcomes other than $x_j$ vanishes, there is no need to maintain any conditional probabilities for other outcomes. In that case, the distribution $\pi_j$ becomes a dimension smaller and is no longer a probability distribution, but rather a likelihood function instead:
$$\pi_j^{new}(\cdot | X_{C(j)}) = L\{X_{C(j)} | X_j = x_j\}$$
$$\propto \Pr\{X_j = x_j | X_{C(j)}\}$$
$$= \pi_j^{old}(x_j | X_{C(j)}).$$
If node k is a successor to node j, $k \in S(j)$, then there is no longer any need for it to depend on the (now certain) value of $X_j$. Therefore the distribution $\pi_k$ also becomes a dimension smaller:
$$\pi_k^{new}(X_k | X_{C^{new}(k)})$$
$$= \pi_k^{old}(X_k | X_{C^{new}(k)}, X_j = x_j),$$
where $C^{new}(k) = C^{old}(k) \setminus \{j\}$, since the arc is now deleted.

Consider the example in Figure 1, in which $N = \{1, 2, 3\}$ and for which $X_2$ is observed. The joint distribution prior to observing $X_2$, displayed in part a, is given by
$$\Pr\{X_N\}$$
$$= \Pr\{X_1\} \cdot \Pr\{X_2 | X_1\} \cdot \Pr\{X_3 | X_2\}$$
$$= \pi_1(X_1) \cdot \pi_2(X_2 | X_1) \cdot \pi_3(X_3 | X_2).$$
The joint distribution posterior to observing $X_2 = x_2$, displayed in part b, follows evidence absorption and is given by:
$$\Pr\{X_N | X_2 = x_2\}$$
$$= \Pr\{X_1\} \cdot L\{X_1 | X_2 = x_2\}$$
$$\cdot \Pr\{X_3 | X_2 = x_2\}$$
$$\propto \pi_1(X_1) \cdot \pi_2^{new}(\cdot | X_1) \cdot \pi_3^{new}(X_3 | \cdot).$$
A simple application of Bayes' Theorem yields
$$\Pr\{X_1 | X_2 = x_2\}$$
$$\propto \Pr\{X_2 = x_2 | X_1\} \Pr\{X_1\}$$
$$\propto L\{X_1 | X_2 = x_2\} \Pr\{X_1\}$$

and $L\{\cdot | X_2 = x_2\} \propto \Pr\{X_2 = x_2\} \propto 1$.
Thus the diagram shown in part b can be transformed into the one shown in part c, in which
$$\pi_1^{new}(X_1) \propto \pi_1^{old}(X_1) \cdot \pi_2^{old}(\cdot | X_1)$$
and $\pi_2^{new}(\cdot) \propto 1$.

This is a modification of arc reversal in which the arc (1, 2) was simply deleted.

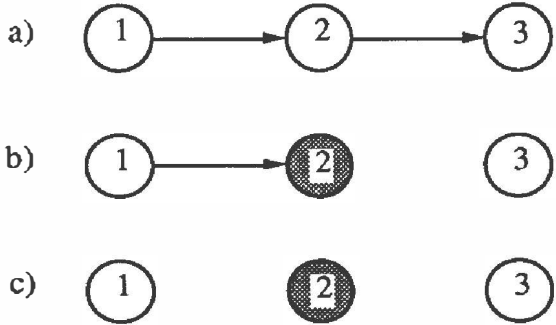

Figure 1. *Simple Example of Evidence Reversal.*

Before extending this example to the general case, care must be taken in interpreting an evidence diagram with observed evidence nodes. In Figure 1, part a, the diagram is drawn prior to the observation of $X_2$. In part b, node 2 is shaded to indicate that its value has been observed. The arc (2, 3) has been deleted to reflect the new distribution at $X_3$ given the observation. A similar change is made for $X_1$ in part c, but it has not yet been performed in part b. When an evidence node is connected to other nodes, their posterior distributions still depend on the likelihood function in the evidence node. Once they become disconnected from the evidence node, their distribution has become posterior to the evidence. The effect of the evidence has not yet been "propagated" through the network until the evidence node is completely disconnected from the other (unobserved) nodes in the diagram. Once it has become disconnected, its likelihood distribution is simply a constant, and it has no effect on the joint distribution for the remainder of the variables. It could be eliminated from the graph, but it is valuable to keep it in the network to indicate that the joint distribution in the network is posterior to the observation.

The general case of Bayes' Theorem with evidence nodes is called <u>evidence reversal</u> (due to its correspondence to arc reversal) and is illustrated in Figure 2. The node $X_i$ conditions the evidence node $X_j$. The other conditional predecessors for the two nodes have been split into three sets corresponding to their common and exclusive conditioning of the nodes. After the evidence reversal operation there is no longer any direct arc between nodes i and j. However, in general, both nodes inherit each other's conditional predecessors.

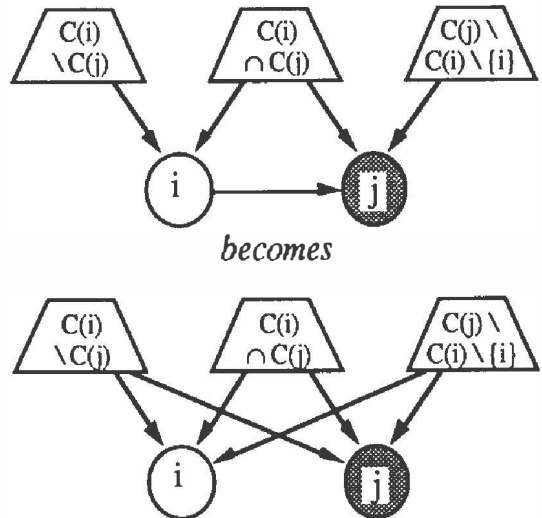

Figure 2. *General Case for Evidence Reversal.*

**Theorem 1. Evidence Reversal.**
*Given an evidence diagram containing an arc from an unobserved probabilistic node i to an evidence node j, but no other directed path from i to j, it is possible to transform the diagram to one with no arc between i and j. In the new diagram, both i and j inherit each other's conditional predecessors.*

Proof:
Let $I = C^{old}(i) \setminus C^{old}(j)$, $J = (C^{old}(j) \setminus C^{old}(i)) \setminus \{i\}$, and $K = C^{old}(i) \cap C^{old}(j)$. The new conditional predecessor sets for i and j are
$$C^{new}(i) = C^{new}(j) = I \cup J \cup K.$$
Because there is no other directed path from i to j, there can be no directed path from i to J, so
$$\Pr\{X_i | X_I X_J X_K\} = \Pr\{X_i | X_I X_K\}$$
$$= \pi_i^{old}(X_i | X_{C^{old}(i)}),$$
$$\Pr\{X_j = x_j | X_i X_I X_J X_K\}$$
$$= \Pr\{X_j = x_j | X_i X_J X_K\}$$
$$\propto L\{X_i X_J X_K | X_j = x_j\}$$
$$= \pi_j^{old}(\cdot | X_{C^{old}(j)}),$$
and
$$\Pr\{X_i, X_j = x_j | X_I X_J X_K\}$$
$$= \Pr\{X_j = x_j | X_i X_I X_J X_K\}$$
$$\cdot \Pr\{X_i | X_I X_J X_K\}$$
$$\propto \pi_j^{old}(\cdot | X_{C^{old}(j)})$$
$$\cdot \pi_i^{old}(X_i | X_{C^{old}(i)})$$

305

$$= \Pr\{X_j = x_j \mid X_I X_J X_K\}$$
$$\cdot \Pr\{X_i \mid X_j = x_j, X_I X_J X_K\}$$
$$\propto \pi_j^{new}(\cdot \mid X_{C^{new}(j)})$$
$$\cdot \pi_i^{new}(X_i \mid X_{C^{new}(i)}).$$

The new likelihood distribution for $X_j$ is found by summing,

$$\pi_j^{new}(\cdot \mid X_{C^{new}(j)})$$
$$= \Sigma_{x_i \in \Omega_i} \pi_j^{old}(\cdot \mid X_{C^{old}(j)})$$
$$\cdot \pi_i^{old}(X_i \mid X_{C^{old}(i)}),$$

and the new conditional probability distribution for $X_i$ is just

$$\pi_i^{new}(X_i \mid X_{C^{new}(i)})$$
$$\propto \pi_j^{old}(\cdot \mid X_{C^{old}(j)})$$
$$\cdot \pi_i^{old}(X_i \mid X_{C^{old}(i)}).$$

The inheritance of conditioning variables is needed to consider the conditional joint distribution for $X_i X_j \mid X_I X_J X_K$. The requirement that there be no other directed (i,j)-path is necessary and sufficient to prevent creation of a cycle. It is equivalent to requiring that there be an ordered list of the nodes in which nodes i and j are adjacent. After the evidence reversal operation that same list will be ordered even if the nodes i and j were switched. #

Once the evidence for a node has been absorbed, the simple operation of evidence reversal can be repeated until there are no more arcs incoming to the evidence node. At that point, it is disconnected from the other nodes and their distributions now reflect the posterior joint distribution. The process whereby all of the arcs into the evidence are deleted is called evidence propagation. It corresponds to the bottom-up propagation in Pearl [1986a] and the first half of global propagation in Lauritzen and Spiegelhalter [1988]. Once the evidence node was absorbed, it no longer had any successors. Therefore, to disconnect it is simply to eliminate any incoming arcs. However, whenever an incoming arc is deleted through evidence reversal, new arcs can be added corresponding to indirect predecessors of the evidence node. This process will continue until the evidence has been reversed with every one of those predecessors. As with most influence diagram reductions, there is considerable freedom in the order in which operations can be performed. In this case, that corresponds to the multiple ways in which the indirect predecessors can appear in an ordered list.

If the diagram is a forest, there is at most one arc into the evidence node at any time, and no arcs are added in the process. Therefore if the diagram is a forest before evidence propagation it will remain so. This is summarized in the following theorem.

**Theorem 2. Evidence Propagation.**
*The evidence from any evidence node j which has been absorbed can be propagated through the belief diagram. First order all of its indirect predecessors. Then perform evidence reversal on the arc from each predecessor to j, in reverse order along the list. Afterwards, node j will be disconnected from the other nodes in the diagram. If the diagram before evidence propagation is a forest, then it will be a forest afterwards. In general, however, if the diagram is a polytree before evidence propagation, it might not be a polytree afterwards.*

An example of evidence propagation on a forest is shown in Figure 3. The value of $X_4$ is observed for the diagram drawn in part a. The diagram after absorption is shown in part b. There are two indirect predecessors for node 4, namely 1 and 2, in that order. Therefore, evidence reversal is first performed on 2 and 4 to obtain the diagram shown in part c, and then it is performed on 1 and 4 to obtain the final diagram, displayed in part d. Notice that the diagram remains a forest, and no new arcs have been created.

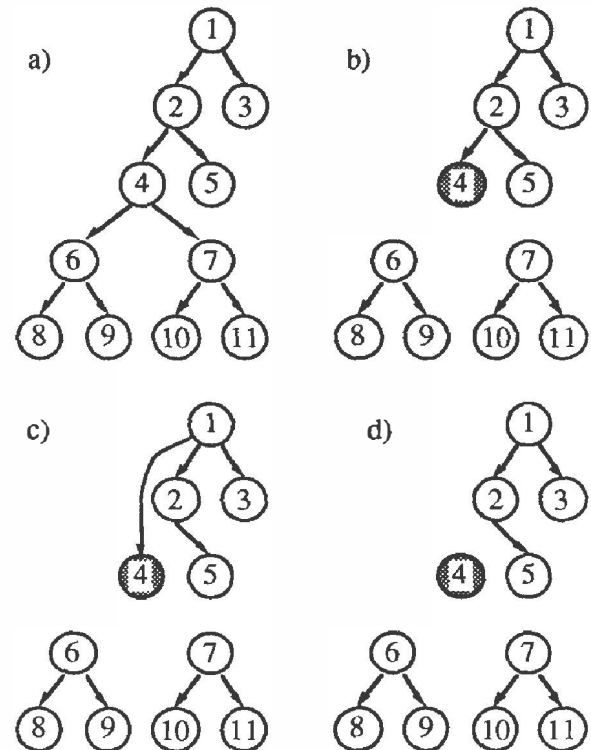

*Figure 3. Evidence Propagation in a Forest.*

Evidence propagation is more complex on a polytree which is not a forest. Such an example is shown in Figure 4. In this case $X_4$ is observed and evidence absorption is performed for node 6 (part a). The indirect predecessors of node 6 are { 1, 2, 3, 4, 5 } and they have multiple orderings, including [ 1 3 2 4 5 ] and [ 2 4 1 3 5 ]. If the first of these is used to guide evidence propagation then evidence reversal is performed for nodes 5 and 6 (part b), 4 and 6 (part c), 2 and 6 (part d), 3 and 6, and finally 1 and 6 (part e). If the second ordering were used then the belief diagram after evidence propagation is shown in part f. Notice that although the diagram was a polytree before evidence propagation, the arcs created in the process made the resulting diagram multiply-connected, and therefore no longer a polytree. Also note that there are multiple representations for the resulting diagram corresponding to different orderings of the evidence reversals.

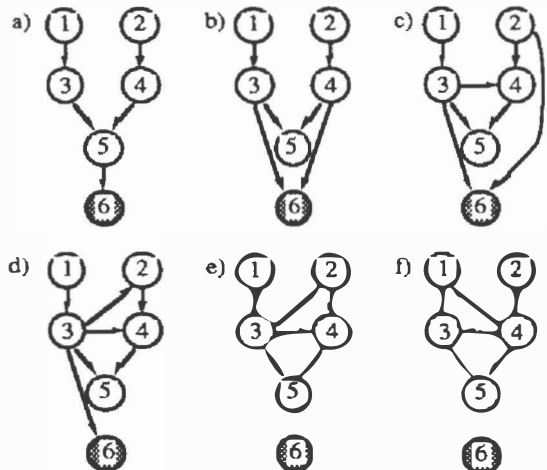

*Figure 4. Evidence Propagation in a Polytree.*

## 4. Probability Propagation

The belief diagram after evidence absorption and propagation represents the posterior joint distribution for the unobserved variables. The operation which computes the marginal distribution for each unobserved variable is called <u>probability propagation</u>. It corresponds to the bottom-down propagation in Pearl [1986a] and the second half of global propagation in Lauritzen and Spiegelhalter [1988].

The fundamental process in probability propagation is conditional expectation. This is a straightforward operation when a network is a polytree, but it can become quite complex in multiply-connected diagrams. As a result, in this paper, the operation will only be described for singly-connected belief diagrams. In general, the process of probability propagation can be viewed as a special case of probabilistic inference as described in Shachter [1988]. When a diagram is multiply-connected, computation of the posterior marginal for some nodes will require changes to the topology of the network, such as arc reversals.

Assuming that the diagram is singly-connected, probability propagation can be performed in a single sweep through the network. This discussion below uses the arc reversal metaphor as in Shachter [1988], but the computations can actually performed without any changes to the network structure. When calculating the posterior marginal distribution $B_j(X_j)$ for node j, all other nodes which are not indirect predecessors of node j are irrelevant (barren) and can be ignored. If node j has no parents (conditional predecessors), then its current distribution is (proportional to) its marginal distribution. If it has parents but no grandparents, then the arcs from j's parents to j can be reversed. Since its parents become irrelevant once their arcs into j are reversed, only the first half of the reversal operation, the node "removal" reduction [Olmsted 1983]) is needed. Thus if node j has parents but no grandparents, its parents can simply be "removed" from the diagram using conditional expectation:

$$B_j(X_j) = \Pr\{X_j\}$$
$$= \sum_{x_{C(j)} \in \Omega_{C(j)}} \Pr\{X_j \mid X_{C(j)} = x_{C(j)}\}$$
$$\cdot \Pr\{X_{C(j)} = x_{C(j)}\}$$
$$= \sum_{x_{C(j)} \in \Omega_{C(j)}} \Pr\{X_j \mid X_{C(j)} = x_{C(j)}\}$$
$$\cdot B_j(x_{C(j)}).$$

At this point node j has no parents and is free to be "removed" into its children. If the diagram is singly-connected, then at every step there will be a node which either has no parents or no grandparents. Unfortunately, if the diagram were multiply-connected then there would be at least one node whose parents would be connected, and for which one of the parents would also be grandparent.

**Theorem 3. Probability Propagation**
*Given a belief diagram in which all evidence has been propagated and which is a polytree, probability propagation can be performed with two simple steps. First, for each node j with no parents (conditional predecessors), set its posterior marginal distribution $B_j(X_j)$ to*

$$B_j(X_j) \propto \pi_j(X_j).$$

*Second, for each node j whose parents all have marginal distributions computed, set*
$$B_j(X_j) = \sum_{x_{C(j)} \in \Omega_{C(j)}} \Pr\{X_j \mid X_{C(j)} = x_{C(j)}\}$$
$$\cdot B_j(x_{C(j)}).$$

*Repeat the second step until posterior marginals have been computed for all nodes.*

Although this simple process will work on any polytree, recall that evidence propagation on a polytree can result in a multiply-connected diagram. In that case, probability propagation would be substantially more complex.

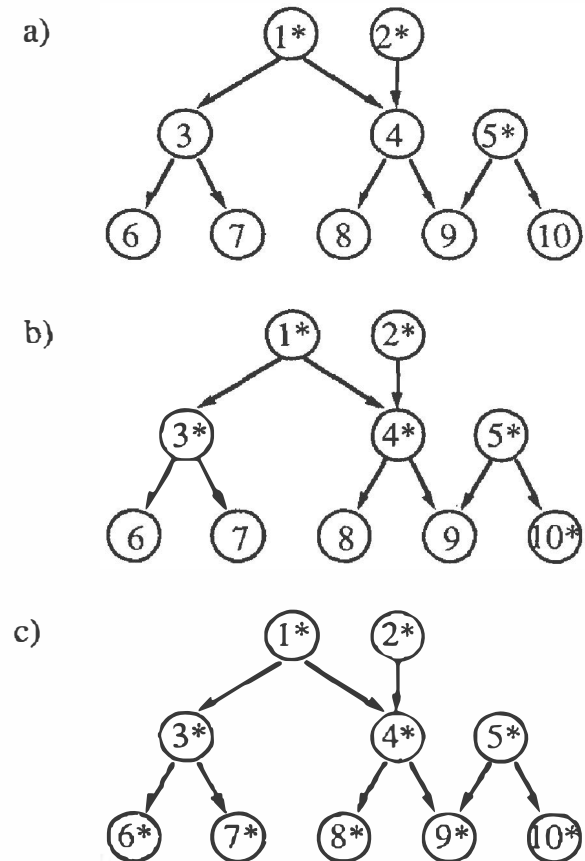

Figure 5. *Probability Propagation in a Polytree.*

An example of probability propagation on a singly-connected network is shown in Figure 5. The nodes labelled with an asterisk ("*") are those for which marginal distributions have been computed. In the first pass (part a) all of the nodes without parents, { 1, 2, 5}, are labelled. In the second pass (part b) those nodes whose parents are all labelled, { 3, 4, 10 } can now be labelled. Finally, in the third pass (part c) the parents are all labelled for all of the remaining nodes, so they too can be labelled and probability propagation is complete.

## 5. Control of the Evidence Process

The three different operations which enable a belief diagram to incorporate experimental evidence into posterior joint and marginals distributions have already been presented: evidence absorption and propagation and probability propagation. In this section, two different control strategies are described to coordinate these operations. Batch processing corresponds to Lauritzen and Spiegelhalter [1988] and message passing corresponds to Pearl [1988]. Finally, it is shown that by instituting message priorities, the two strategies are really the same on a forest.

Although evidence absorption and propagation are fully general operations which can be performed regardless of network topology, probability propagation can only be performed efficiently when the diagram is singly-connected. As has been shown in section 3, if the original diagram was a polytree but not a forest, arcs might be added in evidence propagation which would make the network multiply-connected. If the posterior network is no longer a polytree then probability propagation cannot be performed on the belief diagram. Therefore, in this section when discussing probability propagation, it is assumed that the network stays singly-connected. Nonetheless, the discussion of evidence absorption and propagation apply in general.

There are three different operations in processing experimental evidence. First, when the observations are posted the evidence must be absorbed into the diagram. This operation should always be performed immediately upon the observation of the evidence. Second, the experimental evidence must be propagated through the diagram until the evidence node is disconnected from the rest of the network. This operation can be postponed since the evidence node helps to form the posterior joint distribution even without propagation. Finally, to obtain a posterior marginal distribution at each node, probabilistic propagation is performed. This can be done at any time after the evidence has been propagated.

One method to control these operations is batch processing. Any time after the current evidence has been absorbed, evidence propagation can be conducted. First order all of the nodes in the diagram. Visit each evidence node in reverse order to propagate its evidence. This is done, as explained in section 3, by reversing the arcs to the evidence node from each of its indirect predecessors in reverse order. In fact, however, the arcs to evidence nodes can be reversed in any order provided there is no other directed path from the unobserved node at the tail of the arc to the evidence node at the head of the arc. (There can never be an arc between two absorbed evidence nodes, since evidence absorption eliminates arcs to the successors of evidence nodes.) After the batch processing of evidence propagation, the diagram still contains a posterior joint distribution, but with all of the evidence nodes disconnected from the unobserved nodes.

Whenever the posterior marginal distributions are desired, batch processing can then perform probability propagation (assuming the posterior diagram is a polytree). This can be done in sweeps as in Figure 5, but again the order of operations is not significant: the marginal for any node can be computed once the marginals for all of its parents (if any) have been computed.

Another method to control these operations is through <u>message passing</u>. For a simple description of this method, assume that the initial belief diagram is a forest. In this method the topology of the network does not change, but instead distributions are passed as messages between adjacent nodes for evidence and probability propagation. Instead of drawing an arc (i, j) from unobserved node i to evidence node j and modifying the distribution stored in evidence node j, evidence propagation is performed by sending a message to node i from one of its children, containing the likelihood function which would otherwise be stored in node j,
$$\pi_j(\cdot \mid X_{C(j)}) = \pi_j(\cdot \mid X_i) = L(X_i \mid X_j = x_j).$$
When this message is processed, instead of a new arc being created between i's parent and j, a new likelihood function message is sent to i's parent instead.

Probability propagation can also be performed by sending messages: the newly computed marginal distribution for a node is sent to its children. Whenever the distribution for a node without parents is updated via evidence propagation, that constitutes a new marginal distribution for the node, which should then be sent to its children. Likewise, whenever a node receives a new marginal distribution from its parent, it can then compute a new marginal distribution for itself, and send it to its children.

The probability propagation messages described here are similar but not as efficient as those in Pearl [1986a] since they do not exploit the orthogonality of upward and downward messages. That is, evidence propagation messages necessarily go to all of the indirect predecessors of the evidence node, and then probability propagation messages go to all of their indirect successors. Pearl has shown that any node visited on the way up need not be visited on the way down. However, this redundancy allows for a convenient parallel between message passing and batch processing.

Finally, <u>message priority</u> is a simple control strategy that subsumes batch processing and message passing. Suppose that message passing is being performed on a forest with a priority system with the following four rules:
1. messages of evidence have highest priority;
2. messages from below are always processed before messages from above;
3. any pending message from above is replaced by a subsequent message from above; and
4. if messages are not being received by distributed processors but instead through a single controller, then any message from below should have higher priority than any message from above, and messages from above should be processed in graph order.

This system of message priority has several attractive features. First, if several pieces of evidence are received simultaneously, then this corresponds exactly to batch processing (except when an observed node has children which become disconnected). Second, if a single piece of evidence is received, then this is equivalent to message passing. Finally, in a real-time system this will give immediate attention to evidence absorption, but whenever there is any time for background processing then evidence and probability propagation would be performed.

## 6. Comparisons with the Pearl and the Lauritzen and Spiegelhalter Algorithms

It is well known that the algorithms of Pearl [1986a] and Lauritzen and Spiegelhalter [1988] are closely related. (For example, see the Clustering Algorithm in Pearl [1988].) In this paper, an evidence reversal algorithm was created in the spirit of node reductions and, for diagrams which form a forest, it is equivalent to the other two. In this section we explore some of the consequences of this result, including some insights into how the methods differ once the belief diagram becomes a polytree.

The evidence reversal algorithm always maintains a posterior joint distribution. Since it is essentially the same as the other methods on a forest then they must all be doing the same thing when operating on a forest. This should not be too surprising for the Lauritzen and Spiegelhalter method since it also maintains a posterior joint distribution, albeit through potential functions. However, it is easy to verify that in the case of Pearl's algorithm, adapting his $\lambda$ notation [Pearl 1986a], we can obtain the posterior conditional distributions at any time, using conditional independence and Bayes' Theorem,

$$\Pr\{X_j \mid X_{C(j)}, D_j^+, D_j^-\}$$
$$= \Pr\{X_j \mid X_{C(j)}, D_j^-\}$$

$$\propto L(D_j^- | X_j) \Pr\{X_j | X_{C(j)}\}$$
$$= \lambda_j(X_j) \Pr\{X_j | X_{C(j)}\},$$

where $D_j^+$ is the observed evidence disconnected from node j from above, and $D_j^-$ is the evidence disconnected from below. Another way to describe this is that when the original graph is a forest it is an I-map [Pearl 1988] for the posterior distribution. This is not the case for the polytree, since arcs must be added to the belief diagram during evidence propagation to represent the posterior joint distribution [Pearl 1988].

Although the algorithms are (essentially) identical on a forest, their differences can be illustrated on a polytree. Consider the example from Figure 4. Since the Pearl algorithm does not change the topology of the graph, it can only keep track of the posterior marginals, and not the joint, for a polytree. It computes marginals for nodes 1 and 3 using a factorization of the joint as shown in part e; for nodes 2 and 4 it effectively uses the factorization drawn in part f.

Lauritzen and Spiegelhalter, on the other hand, maintain the posterior joint but not on the original graph. Not only can they adapt the topology to the evidence observed, but in a way that does not always have a directed graph factorization. Consider the evidence diagram from Figure 4 which is redrawn as part a of Figure 6. Although there are many choices for the maximum cardinality ordering used in their algorithm, one form is obtained by first reversing the arc from 2 to 4 (part b), and merging nodes 3 and 4 into a single node (part c). The graph is now a forest, so probability propagation can be performed for any possible evidence.

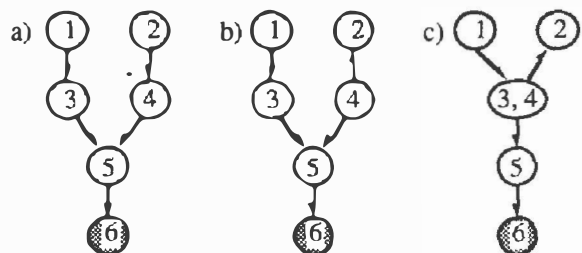

*Figure 6. Lauritzen-Spiegelhalter Algorithm Transforms a Polytree into a Forest.*

Finally, if the evidence reversal method were applied to the polytree diagram drawn in part a of Figure 6 (rather than its forest variant drawn in part c), then the evidence could be propagated without any problem (as shown in Figure 4), but the posterior belief diagram would be multiply-connected, making probability propagation a substantially more complex task.

## 7. Conclusions

The evidence reversal operation on the belief diagram extends the arc reversal and node reduction metaphor on general influence diagrams with decisions to variables whose outcomes are observed prior to the earliest decision.

The close relationship between this method and the methods of Pearl [1986a] and Lauritzen and Spiegelhalter [1988] provide some additional insights which should prove useful in teaching and extending the state of the art techniques for exact analysis of probabilistic inference problems. It gives some feeling for the elegance of Pearl's method applied to polytrees and the general power of the Lauritzen and Spiegelhalter approach. Finally, the connection between batch processing and message passing control for the evidence reversal algorithm suggests simple yet efficient ways to combine the powers of all three methods.

## References

Dempster, A. P. and A. Kong. 1986. Uncertain Evidence and Artificial Analysis, Research Report S-108, Department of Statistics, Harvard University.

Howard, R. A., and J. E. Matheson. 1981. Influence Diagrams. In The Principles and Applications of Decision Analysis, Vol II, (1984), R. A. Howard and J. E. Matheson (eds.). Strategic Decisions Group, Menlo Park, Calif.

Lauritzen, S. L., and D. J. Spiegelhalter. 1988. Local Computations with Probabilities on Graphical Structures and their Application to Expert Systems (with Discussion), JRSS B, 50, 157-224.

Olmsted, S. M. 1983. On Representing and Solving Decision Problems. Ph.D. Thesis, EES Dept., Stanford University.

Pearl, J. 1986a. Fusion, Propagation and Structuring in Belief Networks. AIJ, 29:241-288.

Pearl, J. 1986b. A Constraint-Propagation Approach to Probabilistic Reasoning, in L. N. Kanal and J. F. Lemmer (Editors), Uncertainty in Artificial Intelligence, North-Holland, 357-370.

Pearl, J. 1988. Probabilistic Reasoning in Intelligent Systems. Morgan Kaufmann, San Mateo, Calif.

Shachter, R. D. 1986. Evaluating Influence Diagrams. Opns Rsch 34, 871-882.

Shachter, R. D. 1988. Probabilistic Inference and Influence Diagrams. Opns Rsch 36, 589-604.

Shafer, G., P. P. Shenoy, P. P., and L. Mellouli. 1987. Propagating Belief Functions in Qualitative Markov Trees, IJAR, 1, 349-400.